\documentclass[10pt,twocolumn,letterpaper]{article}

\usepackage[pagenumbers]{cvpr} % To force page numbers, e.g. for an arXiv version

\usepackage{graphicx}
\usepackage{amsmath}
\usepackage{amssymb}
\usepackage{booktabs}
\usepackage[accsupp]{axessibility}

\usepackage[pagebackref,breaklinks,colorlinks]{hyperref}

\usepackage[capitalize]{cleveref}
\crefname{section}{Sec.}{Secs.}
\Crefname{section}{Section}{Sections}
\Crefname{table}{Table}{Tables}
\crefname{table}{Tab.}{Tabs.}

\def\cvprPaperID{10928} % *** Enter the CVPR Paper ID here
\def\confName{CVPR}
\def\confYear{2023}

\begin{document}

\title{SMPConv: Self-moving Point Representations for Continuous Convolution}

\newcommand\CorrespondingAuthorMark{\footnotemark[\arabic{footnote}]}
\author{Sanghyeon Kim$^1$ \ \ \ \ \ \ Eunbyung Park$^{1,2}$\thanks{Corresponding authors} \\
$^1$Department of Electrical and Computer Engineering, Sungkyunkwan University \\
$^2$Department of Artificial Intelligence, Sungkyunkwan University
% {\tt\small \{shkim960520, epark\}@skku.edu}
% \and
% Eunbyung Park\\
% Department of Artificial Intelligence\\
% % First line of institution2 address\\
% % {\tt\small secondauthor@i2.org}
}
\maketitle

%%%%%%%%% ABSTRACT
\begin{abstract}
Continuous convolution has recently gained prominence due to its ability to handle irregularly sampled data and model long-term dependency. Also, the promising experimental results of using large convolutional kernels have catalyzed the development of continuous convolution since they can construct large kernels very efficiently. Leveraging neural networks, more specifically multilayer perceptrons (MLPs), is by far the most prevalent approach to implementing continuous convolution. However, there are a few drawbacks, such as high computational costs, complex hyperparameter tuning, and limited descriptive power of filters.
% First, it introduces significant computational costs to the process of training deep convolutional neural networks (CNNs). This has thus far restricted its broader adoption, and they have yet to be tested in large-scale experiments. Second, since it uses neural networks to generate kernels, more hyperparameters are added to the already-complicated machine learning development process. Lastly, the architectural priors implicitly induced by the kernel-generating networks often limit the descriptive power of the learned filters.
This paper suggests an alternative approach to building a continuous convolution without neural networks, resulting in more computationally efficient and improved performance. We present self-moving point representations where weight parameters freely move, and interpolation schemes are used to implement continuous functions. When applied to construct convolutional kernels, the experimental results have shown improved performance with drop-in replacement in the existing frameworks. Due to its lightweight structure, we are first to demonstrate the effectiveness of continuous convolution in a large-scale setting, e.g., ImageNet, presenting the improvements over the prior arts. Our code is available on \href{https://github.com/sangnekim/SMPConv}{https://github.com/sangnekim/SMPConv}
\end{abstract}

%%%%%%%%% BODY TEXT
\section{Introduction}
\label{sec:intro}

There has been a recent surge of interest in representing the convolutional kernel as a function over a continuous input domain. It can easily handle irregularly sampled data both in time~\cite{warden2018speech, bagnall2018uea} and space~\cite{wu20153d, yi2016scalable}, overcoming the drawbacks of the discrete convolution operating only on discretized sampled data with pre-defined resolutions and grids. With the progress in modeling and training continuous kernels, it has enjoyed great success in many practical scenarios, such as 3D point cloud classification and segmentation~\cite{wang2018deep, xu2018spidercnn, Pointnet, interpconv, thomas2019kpconv}, image super resolution~\cite{vasconcelos2022cuf}, object tracking~\cite{danelljan2016beyond}, to name a few. Furthermore, the recent trends of using large convolutional kernels with strong empirical results urge us to develop a more efficient way to implement it~\cite{replknet, slak}, and the continuous convolution will be a promising candidate because of its capability to readily construct arbitrarily large receptive fields~\cite{romero2021ckconv, romero2021flexconv}. 

One of the dominant approaches to modeling the continuous kernel is to use a particular type of neural network architecture, taking as inputs low-dimensional input coordinates and generating the kernel values~\cite{romero2021ckconv, romero2021flexconv}, known as neural fields~\cite{sitzmann2020implicit, mildenhall2020nerf} or simply MLPs. Using neural fields to represent the kernels, we can query kernel values at arbitrary resolutions in parallel and construct the large kernels with a fixed parameter budget, as opposed to the conventional discrete convolutions requiring more parameters to enlarge receptive fields. Thanks to recent advances to overcome the spectral bias on training neural fields~\cite{sitzmann2020implicit}, they can also represent functions with high-frequency components, which enables learned kernels to capture fine details of input data. 

While promising in various tasks and applications, this approach has a few downsides. First, it incurs considerable computational burdens to already computation-heavy processes of training deep neural networks. Each training iteration involves multiple forward and backward passes of MLPs to generate kernels and update the parameters of MLPs. This additional complexity prevents it from being applied to large-scale problems, such as ImageNet-scale, since it needs deeper and wider MLP architectures to construct more complex kernels with more input and output channels. Although MLPs can generate larger sizes and numbers of kernels without adding more parameters, it has been known that the size of MLPs mainly determines the complexity of the functions they represent and, eventually, the performance of the CNNs.

Furthermore, the kernels generated by an MLP depend heavily on the architectural priors. As a universal function approximator, a neural network with sufficient depth and width can express any continuous functions~\cite{hornik1989multilayer}. However, we have empirically observed strong evidence that the architecture of neural networks has played a significant role in many practical settings, suggesting that various architectural changes to the kernel-generating MLPs would significantly affect the performance of CNNs. Considering a large number of hyperparameters in training neural networks, adding more knobs to tune, e.g., activation functions, width, depth, and many architectural variations of MLPs., would not be a pragmatic solution for both machine learning researchers and practitioners. 

In this paper, we aim to build continuous convolution kernels with negligible computational cost and minimal architectural priors. We propose to use moving point representations and implement infinite resolution by interpolating nearby moving points at arbitrary query locations. The moving points are the actual kernel parameters, and we connect the neighboring points to build continuous kernels. Recent techniques in neural fields literature inspire it, where they used grid or irregularly distributed points to represent features or quantities in questions (density or colors) for novel view synthesis~\cite{sun2022dvgo, yu2021plenoxels, chen2022tensorf, pointnerf1, pointnerf2}. The suggested approach only introduces minor computational costs (interpolation costs) and does not consist of neural networks (only point representations and interpolation kernels). Moreover, the spectral bias presented in training MLPs~\cite{rahaman2019spectral} does not exist in the suggested representation. Each point representation covers the local area of the input domain and is updated independently of each other, contrasted with MLPs, where updating each parameter would affect the entire input domain. Therefore, highly different values of nearby points can easily express high-frequency components of the function. 

The proposed method can also be more parameter efficient than the discrete convolution to construct the large kernels. Depending on the complexity of the kernels to be learned, a few numbers of points may be sufficient to cover a large receptive field (e.g., a unimodal function can be approximated by a single point). Many works have extensively exploited non-full ranks of the learned kernels to implement efficient convolutions or compress the models~\cite{slak}. Our approach can likewise benefit from the presence of learned kernels with low-frequency components.

We conduct comprehensive experimental results to show the effectiveness of the proposed method. First, we demonstrate that moving point representations can approximate continuous functions very well in 1D and 2D function fitting experiments. Then, we test its ability to handle long-term dependencies on various sequential datasets. Finally, we also evaluate our model on 2D image data. Especially, we perform large-scale experiments on the ImageNet classification dataset (to our knowledge, it is the first attempt to use continuous convolution for such a large-scale problem). Our proposed method can be used as a drop-in replacement of convolution layers for all the above tasks without bells and whistles. The experimental results show that it consistently improves the performance of the prior arts.

%-------------------------------------------------------------------------
\section{Related works}
\begin{figure*}[ht]
% \vspace{-.5cm}
  \centering
  \includegraphics[width=\textwidth]{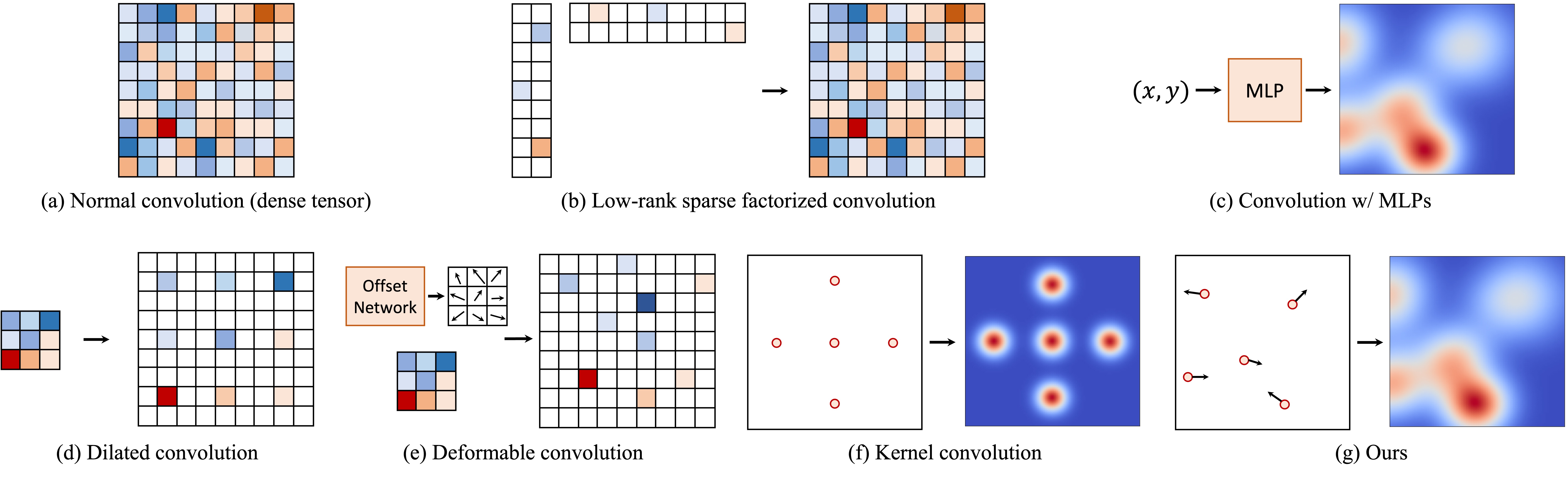}
  \caption{A various methods for large kernel construction.}
  \label{fig:large_kernels}
  \vspace{-0.5cm}
\end{figure*}

\textbf{Neural fields and continuous convolution.}
Neural fields have recently emerged as an alternative neural network representation~\cite{sitzmann2020implicit}. It is a field parameterized by a neural network, such as simple MLP, taking as low-dimensional coordinates input and generating quantity in questions. It has shown great success in various visual computing domains, such as novel view synthesis~\cite{mildenhall2020nerf}, 3D shape representation~\cite{mescheder2019occupancy,park2019deepsdf}, and data compression~\cite{dupont2021coin}, to name a few. Since it is a function over a continuous input domain, it can produce outputs at arbitrary resolutions. Recent studies have exploited its continuous nature to model continuous kernels for CNNs~\cite{wang2018deep, romero2021ckconv, romero2021flexconv}. \cite{wang2018deep} used an MLP architecture to implement the continuous kernel to handle irregularly sampled data, such as 3D point clouds. While successful, the descriptive power of the learned kernel is limited due to its bias toward learning low-frequency components. With the recent advances in overcoming the spectral bias~\cite{sitzmann2020implicit}, \cite{romero2021ckconv} has explored various activation functions to improve the performance of CNNs. To further improve, \cite{romero2021flexconv} proposed to learn the receptive field sizes and showed impressive performance on multiple tasks. 

We also propose a method that can likewise be classified as a neural field. However, we implement a field without neural networks, using self-moving point representations and interpolation schemes. It significantly reduces the computational costs to compute the kernels during training compared to the conventional MLP-based methods. Furthermore, it removes the burden of numerous hyperparameter searches for a newly introduced neural network in the already complicated training procedure.

\textbf{Grid and point representations.}
Although MLP-based neural fields have succeeded in many tasks, they require substantial computational costs for training and inference, and the spectral bias presented in MLPs often degrades the performance. In order to reduce the computations and avoid the issues of using MLPs, classical grid-based representations have been adopted in the neural fields. Plenoxels~\cite{yu2021plenoxels} stores the coefficients of spherical harmonic in the 3D voxel structure and implements the infinite resolution via the interpolation methods, reducing the training time from days to minutes. Instant-NGP~\cite{muller2022instant} further optimized the speed using a hash-based indexing algorithm. The combination of grid representations and MLPs has been extensively explored to find better solutions~\cite{sun2022dvgo, chen2022tensorf}.

Point representations have been recently suggested in neural fields to improve the shortcomings in grid-based representations, thanks to their flexibility and expressibility~\cite{pointnerf1,pointnerf2}. They can generate outputs over a continuous input domain by interpolating neighboring points given a query point. It can be more accurate and parameter efficient than the grid-based ones since it can adaptively locate points, considering the complexity of functions, e.g., fewer points in low-frequency areas. Although our work is largely inspired by these works, 1. we repurpose the point representations to build continuous convolutional kernels, 2. we do not use MLPs that would introduce additional modeling complexity, 3. our method does not require complex initializations (or algorithms) to relocate points, such as depth-based initialization~\cite{pointnerf1} and mask-based initialization~\cite{pointnerf2}.

\textbf{Large kernel convolution.}
Since the great success of VGG-style CNNs~\cite{simonyan2014vgg, he2016resnet, howard2017mobilenets, tan2019efficientnet} in ImageNet~\cite{deng2009imagenet}, we repeatedly apply small kernels (e.g., $3 \times 3$ kernel) with deep layers to get a large receptive field and deal with long-term dependencies. With the massive success of transformers in the vision domain~\cite{dosovitskiy2020vit, touvron2021deit,liu2021swin,dong2022cswin}, which have a large receptive field, recent studies have started to revisit the descriptive power of large kernels~\cite{replknet, slak, tomen2021spectral, liu2022convnet}. RepLKNet~\cite{replknet} has shown we can increase the kernel size up to $31 \times 31$ with the help of the reparameterization trick. More recently, SLaK~\cite{slak} managed to scale the kernel size to $51 \times 51 $ and $61 \times 61$, showing improved performance with dynamic sparsity and low-rank factorization (\cref{fig:large_kernels}-(b)). While promising, the number of parameters increases according to the kernel size, which is a major bottleneck in representing large kernels. 
% While promising, due to discrete nature, it is limited to representing extremely large kernels.

The continuous kernel can construct large kernels with a fixed parameter budget (\cref{fig:large_kernels}-(c)). CKConv~\cite{romero2021ckconv} and FlexConv~\cite{romero2021flexconv} have exploited this property and demonstrated that their method can model long-term dependencies by constructing large kernels on various datasets. However, they introduce a considerable amount of computations to the training process, and they have yet to perform large-scale experiments, e.g., ImageNet. To the best of our knowledge, our approach is the first to conduct such a large-scale experiment using continuous convolution.

On the other hand, dilated convolutions~\cite{chen2017deeplab, chen2017deeplab2} can also be used to enlarge receptive fields with a small number of parameters, and they also do not introduce additional computations during training (\cref{fig:large_kernels}-(d)). Deformable convolutions~\cite{dai2017deformable} look similar to ours in terms of moving points arbitrarily (\cref{fig:large_kernels}-(e)). However, they learn how to deform the kernels (or predict the offset) during inference. On the other hand, we adjust the locations of the point representations during training to find the optimal large kernels. A concurrent work suggests learning the offsets during training~\cite{hassani2021spacings}. In contrast to ours, it is a discrete formulation, thus losing the benefits of continuous convolution. Furthermore, we adjust the receptive fields of each point representation separately, yielding more expressive representations.

\textbf{Continuous convolution for point clouds.}
There have been many continuous convolution approaches to handle 3D point cloud data, which is an important example of irregularly sampled data. PointNet~\cite{Pointnet} and PointNet++~\cite{qi2017pointnet++} are pioneer works that use average pooling and $1 \times 1$ convolution to aggregate features. ~\cite{wang2018deep, wu2019pointconv, li2018pointcnn, liu2019relation} leveraged MLPs to implement continuous convolution. KPConv~\cite{thomas2019kpconv} is also considered a continuous convolution and shares some similarities with ours (\cref{fig:large_kernels}-f). They also used point representation and interpolation kernels for handling point clouds. However, their points are fixed over the training, unlike ours. They also proposed a deformable version, which requires additional neural networks to predict the offset of the kernel points. 

\section{SMPConv}
% In this section, we describe the proposed method. First, we introduce self-moving point representations, which is capable of representing continuous functions. Second, we explain how to apply the suggested representation to construct the convolution operator. And, then we briefly discuss the method and large kernel construction. Finally, we discuss training the SMPConv.

\subsection{Self-moving point representation}
This section describes the proposed self-moving point representation to represent a continuous function. Let $d$ be the size of the input coordinates dimension, e.g., $1$ in time-series data and $2$ in the spatial domain. $\texttt{SMP}: \mathbb{R}^d \rightarrow \mathbb{R}^{N_c}$ is a vector-valued function, mapping from the input coordinates to the output kernel vectors, where $N_c$ is a channel size. Given a query point $x \in \mathbb{R}^d$, we define a continuous kernel function as follows.
\begin{equation}
    \texttt{SMP}(x; \phi) = \frac{1}{|\mathcal{N}(x)|}\sum_{i \in \mathcal{N}(x)}g(x, p_i, r_i)w_i,
\end{equation}
where $\phi=\{\{p_i\}_{i=1}^{N_p}, \{w_i\}_{i=1}^{N_p}, \{r_i\}_{i=1}^{N_p} \}$ is a set of learnable parameters, and $N_p$ is the number of points that are used to represent the function. $p_i \in \mathbb{R}^d$ is the coordinates of self-moving point representation $w_i \in \mathbb{R}^{N_c}$, and each point representation also has a learnable radius, $r_i \in \mathbb{R}^+$ is a positive real number, which we implement it by clipping for numerical stability. We define a distance function $g: \mathbb{R}^{d} \times \mathbb{R}^{d} \times \mathbb{R}^{+} \rightarrow \mathbb{R}$ as follows,
\begin{equation}\label{eq:distance_function}
     g(x, p_i, r_i) = 1- \frac{\Vert x-p_i \Vert_1}{r_i},
\end{equation}
 where $\Vert \cdot \Vert_1$ is a L1 distance. $\mathcal{N}(x)$ is a set of indices of neighboring points of a query coordinate $x$, defined as $\mathcal{N}(x) = \{i\,|\,g(x, p_i, r_i)>0,\,\forall i\}$.
% For numerical stability, we added a small number $\epsilon$ in the denominator, which was set to $=0.0001$ throughout the paper. 
Thus, points beyond a certain distance (depending on the radius) will not affect the query point. Hence, $\texttt{SMP}$ generates output vectors by a weighted average of the nearby point representations. Note that all three parameters $\{p_i\}, \{w_i\}, \{r_i\}$ are jointly trained with the CNN model parameters, and the gradients w.r.t those parameters can be easily computed using an automatic differentiation library. As the name $\texttt{SMP}$ suggests, the coordinates $\{p_i\}$ are updated during training, resulting in moving points representation.

Compared to a fixed-point representation, where $\{p_i\}$ are not trainable, ours can approximate complex functions more precisely. Since each point can move freely, more points can be gathered in high-frequency areas. On the other hand, few points can easily represent low-frequency components, resulting in more parameter-efficient representation. For example, a single point may be sufficient to approximate unimodal functions. 

\subsection{SMPConv}
\begin{figure}[t]
  \centering
  \includegraphics[width=0.45\textwidth]{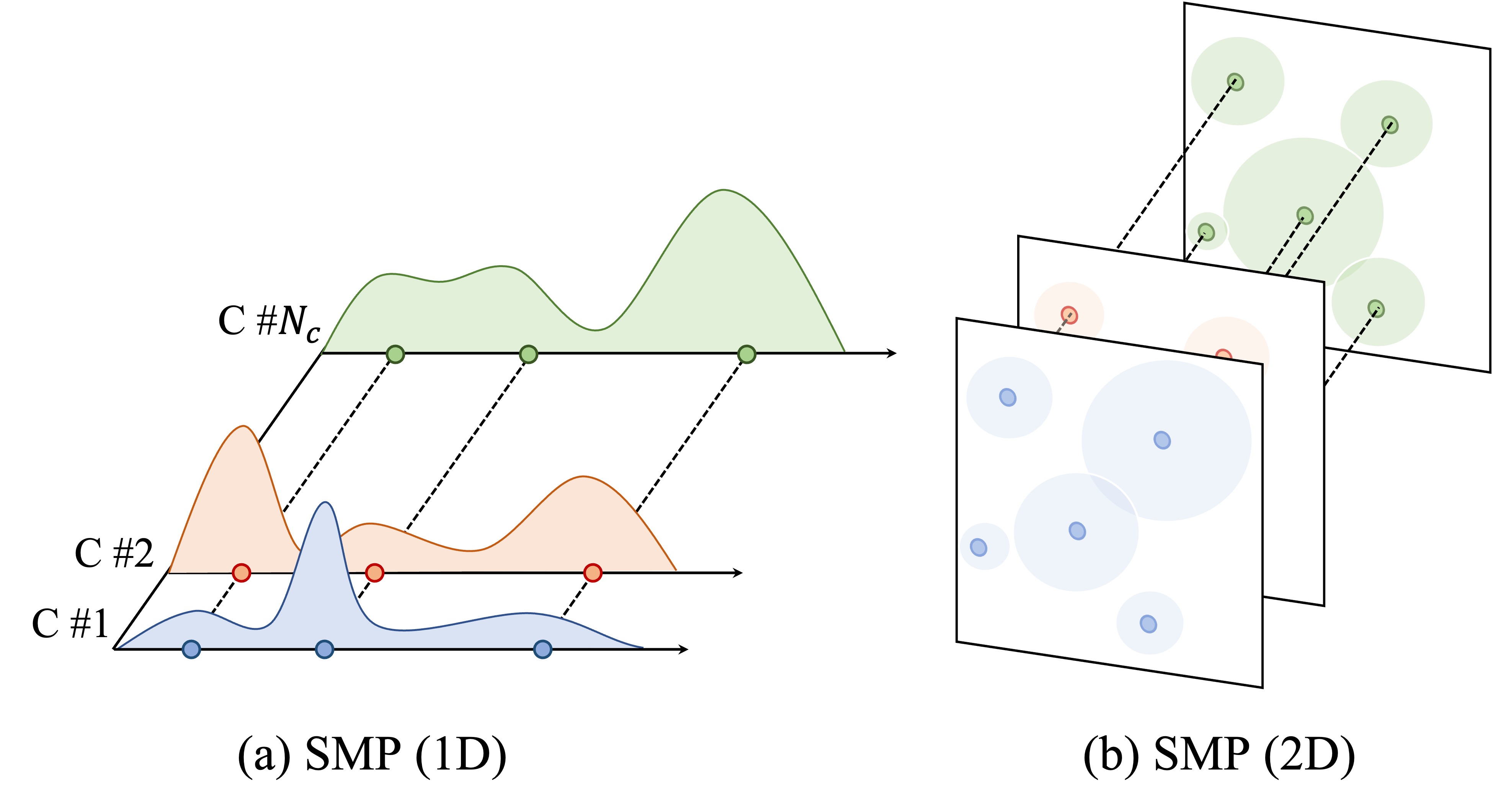}
  \caption{Self-moving point representation. (a) SMP as a function of the one-dimensional input domain, and (b) SMP as a function of the two-dimensional input domain. `C \#1' means the first channel. Each channel shares the location of the points, whereas each channel has its own weight parameters.}
  \label{fig:smp}
\end{figure}
We leverage the suggested representation to implement a continuous convolution operator. In one dimensional case, $d=1$, a continuous convolution can be formulated as,
\begin{equation}
     (f \ast \texttt{SMP})(x) = \sum_{c=1}^{N_c} \int_\mathbb{R} f_c(\tau)\texttt{SMP}_c(x-\tau)d\tau,
\end{equation}
where $f: \mathbb{R} \rightarrow \mathbb{R}^{N_c}$ is a input function and the $f_c$ and $\texttt{SMP}_c$ denote the $c$-th element of the inputs. The convolution operator generates a function, computing the filter responses by summing over entire $N_c$ channels. One $\texttt{SMP}$ representation corresponds a convolution operator, and multiple $\texttt{SMP}$s are used to implement one convolutional layer to generate multiple output channels. In contrast to the previous MLP-based continuous convolution, which uses one neural network for one convolutional layer, our approach has separate parameters for each convolution filter in a layer. It gives more freedom to each filter and results in more descriptive power of the learned filter. 

As depicted in \cref{fig:smp}, the kernels of each filter share the position parameters. That is, each filter of one layer has its own position parameters. Although we could use different $\texttt{SMP}$ for different channels. However, it will considerably increase the number of learnable parameters ($\{p_i\}$ per channel), and we believe that locating points at the same location for a convolutional filter can be a reasonable prior, where a convolutional filter can focus on specific areas in the input domain.

We leverage our continuous formulation to construct large kernels, motivated by the recent success of using them in many tasks. We can create arbitrary size large kernels by querying multiple discretized coordinates to $\texttt{SMP}$.

% \subsection{Large kernel construction}
% We leverage our continuous formulation to construct large kernels, motivated by the recent success of using them in many tasks. We can create arbitrary size kernels with a fixed parameter budget by querying multiple discretized coordinates to $\texttt{SMP}$. A typical CNN architecture using our method consists of two types of parameters, one generated by ours and the other parameters, such as ones in $1 \times 1$ convolutional layer. Both parameters can be jointly trained using standard gradient-based optimization algorithms.

\subsection{Training}
Training a large kernel has been challenging and computationally heavy, and naive training practice has yet to show promising results. Recently, \cite{ding2021repvgg, replknet} proposed a reparameterization trick to combine different-size kernels as a separate branch, resulting in improved performance and more stable training. We also applied the same trick to train CNNs with SMPConv. 

We empirically figured out that performance degradation occurs when the coordinates $\{p_i\}$ are forced to fit inside the kernel by clipping. Thus, we let the coordinates be freely updated during training.

We also found that the initialization of the parameters $\phi$ matters. For point locations, $\{p_i\}$, we randomly sample from a gaussian distribution with small $\sigma$. It initially locates the points in the center and gradually spreads out over the training process. We empirically found that this strategy yields more stable training, especially at the beginning of the training. We also initialized with small values for $\{r_i\}$, which each weight parameter firstly has a narrow sight. Over the course of training, $\{r_i\}$ also gradually increases if necessary.

\subsection{Efficiency}
\begin{table}[t]
\centering
\resizebox{0.45\textwidth}{!}{
    \begin{tabular}{c|c|c|c|c}
        \noalign{\smallskip}\noalign{\smallskip}\hline
        Method & $k$ & Params. & Time $\downarrow$ & Throughput $\uparrow$\\
        \hline
        & 3 & 0.29M  & 61.2 & \textbf{4390.7}\\
        \cline{2-5}
        Deformable~\cite{dai2017deformable} & 5 & 1.37M & 157.3 & 1618.9\\
        \cline{2-5}
        & 7 & 4.39M & 293.1 & 882.3\\
        \hline
        FlexConv~\cite{romero2021flexconv} & 33 & 0.67M & 92.9 & 1923.4\\
        \hline
        SMPConv& 33 & 0.49M & \textbf{40.1} & 4258.4\\
        \hline
    \end{tabular}}
\caption{Training time (sec/epoch) and throughput (examples/sec) comparison with CIFAR10 on a single RTX3090 GPU. Both are tested with a batch size of 64 and input resolution of 32 $\times$ 32. The $k$ is kernel size. The time is the average training time of the first 10 epochs.}
\label{table:deformable}
\end{table}

Assuming the size of a convolution filter is $C \times N \times N$, where $C$ is the number of kernels and $N$ is the height and width of filter, $CN^2$ parameters are required in dense convolution (\cref{fig:large_kernels}-(a)). Therefore, the number of parameters is proportional to kernel resolution $N \times N$. On the other hand, $\texttt{SMP}$ needs $(1+d+C)N_p$ parameters, where $d$ is the size of the input coordinates dimension, and $N_p$ is the number of weight points. We used $N_p\ll N^2$, so $\texttt{SMP}$ is more efficient than dense convolution in terms of the number of parameters. Furthermore, as the number of parameters does not depend on kernel resolution, $\texttt{SMP}$ can represent kernels of any size, such as large or continuous kernels with fixed budget parameters.

Due to the point representations and interpolation schemes without supplementary neural networks, SMPConv has an advantage of computational complexity. On the other hand, the existing large kernel convolutions are computationally heavy. Deformable convolution (\cref{fig:large_kernels}-(e)), for example, requires offset prediction networks and convolution with interpolated inputs \textit{during both training and inference}, resulting in additional computation and parameter costs. Additionally, it relies on dense convolution, making it impractical to increase kernel size significantly. Similarly, MLP-based methods(\cref{fig:large_kernels}-(c)) like FlexConv~\cite{romero2021flexconv} leverage kernel generation neural network, and it also increases computational burdens. The results presented in \cref{table:deformable} demonstrate that SMPConv outperforms the existing large kernel convolutions in terms of speed.

\section{Experiments}
% In this section, we have performed a variety of experiments to analyze and compare the performance of our proposed method. First of all, we presented a study that shows that the proposed representation can approximate arbitrary continuous functions. Secondly, we evaluated our method on various sequential data classification tasks to demonstrate our method's capability to model long-term dependencies. Next, we tested our method on 2D image classification tasks in the large kernel domain. We also performed large-scale experiments on the 1K ImageNet classification task, comparing our method to the current state-of-the-art methods. A few ablation studies have also been provided to justify the choice of our method. Overall, we performed an in-depth analysis, presented visualizations, and provided a detailed discussion of our proposed method. 

\subsection{Continuous function approximation}
\begin{figure}[t]
  \centering
  \includegraphics[width=0.45\textwidth]{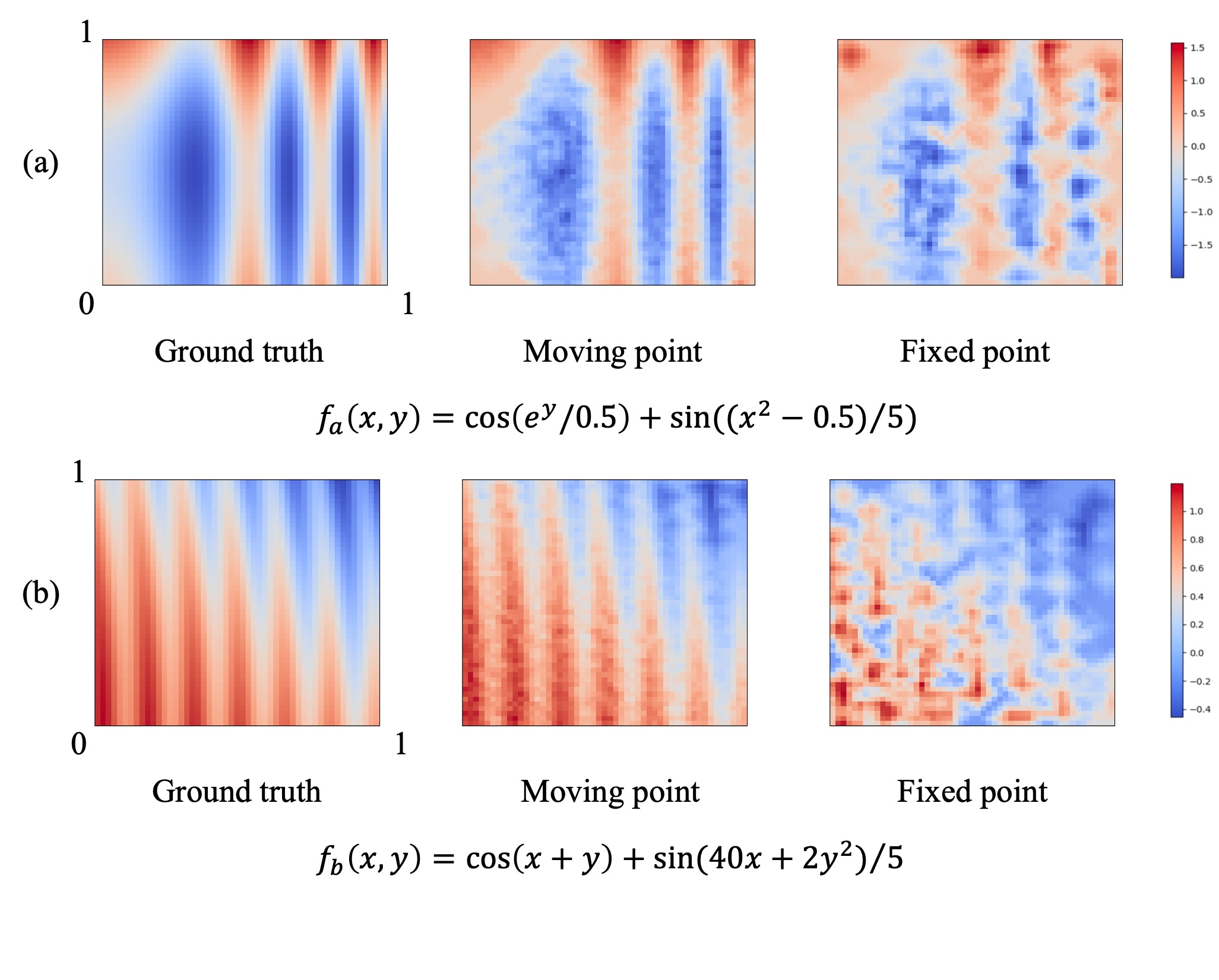}
  \vspace{-.5cm}
  \caption{Comparison between the moving point and the fixed point representations through the fitting. Our proposed moving point representations can approximate given continuous functions with higher accuracy.}
  \label{fig:functionfit}
\end{figure}

Firstly, we conducted a fitting experiment to validate that our self-moving point representation can work as an approximator for a continuous function. To do so, we used two sinusoidal-based functions as the ground truth. Given a function, $\texttt{SMP}$ is optimized to represent the sampled function on a $51 \times 51$ grid. In this experiment, we designed $\texttt{SMP}$ with 204 points. The fitting result has been shown in \cref{fig:functionfit}. It demonstrates that our proposed method reasonably well approximates a given continuous function with fewer points. Additionally, we compared with the fixed point representation and observed that optimizing the position of points together helps to better approximate the function as the number of points is equal.
% We have implemented Eq.~(\ref{eq:fa}) and Eq.~(\ref{eq:fb}) in experiments denoted by ~\cref{fig:functionfit} (a) and ~\cref{fig:functionfit} (b), respectively.

% \begin{equation}\label{eq:fa}
%     f(x,y) = \text{cos}(\frac{e^y}{0.5}) + \text{sin}(\frac{x^2 - 0.5}{5})
% \end{equation}

% \begin{equation}\label{eq:fb}
%     f(x,y) = \text{cos}(x+y) + \frac{\text{sin}(40x + 2y^2)}{5}    
% \end{equation}

\subsection{Sequential data classification}
To demonstrate that SMPConv can handle long-term dependencies well, we evaluated our method on various sequential data tasks, such as sequential image and time-series classification. To do so, we followed FlexTCN ~\cite{romero2021flexconv} to construct a SMPConv architecture for causal 1D CNN whose kernel size is same as the input sequence length. We substituted their parameterized kernels with ours without additional modifications. To maintain a similar number of network parameters, SMPConv contains 30 weight points in each $\texttt{SMP}$. To alleviate the computation burdens caused by the convolution with large kernels, we have considered exploiting the computations through a fast Fourier transform. More network and experimental details are in \cref{appxsec1.1}.
% For this, the data is converted to the Fourier domain and multiplication is applied instead of convolution, and then the result is obtained through the inverse Fourier transform. 

% \subsubsection{Sequential image}
\textbf{Sequential image.}
We tested our SMPConv on the 1D version of images from the datasets, sequential MNIST (sMNIST), permuted MNIST(pMNIST)~\cite{le2015simple}, and sequential CIFAR10 (sCIFAR10)~\cite{chang2017dilated}. These datasets have long input sequence lengths, for example, 784 for sMNIST and pMNIST, and 1024 for sCIFAR10. Note that it is hard to model these datasets without proper kernel representations. As shown in \cref{table:seqimage}, the proposed model has achieved state-of-the-art results on both sMNIST and pMNIST. For sCIFAR10 dataset, our model has outperformed all the comparative models except S4\cite{gu2021s4}. Compared with the FlexTCN, which has a similar network base, our model improved the accuracy by 4$\%$. These results show that our proposed model is suitable and effective for sequential images.

% \subsubsection{Time-series}
\textbf{Time-series.}
We evaluated our model on time-series sequence datasets, character trajectories (CT)~\cite{bagnall2018uea}, and speech commands (SC)~\cite{warden2018speech}. The results have been displayed in \cref{table:timeseires}. In the relatively shorter MFCC features data, SMPConv achieved test accuracy similar to FlexTCN. To validate that our proposed model can model extremely long sequences, we conducted experiments on the SC-raw dataset, which has a sequence length of 16000. Similar to the sequence image classification result, our model outperformed FlexTCN with a large margin of +3$\%$.

Compared to other models, our SMPConv has achieved considerably better performance for both sequential image and time-series classification. It ensures that our kernel representation is capable of handling long-term dependencies even in the case of a limited number of parameters.

\begin{table}
\centering
\resizebox{0.45\textwidth}{!}{
    \begin{tabular}{c c|c|c|c}
        \noalign{\smallskip}\noalign{\smallskip}\hline
        Model & Params. & sMNIST & pMNIST & sCIFAR10 \\
        \hline
        DilRNN~\cite{chang2017dilated} & 44k & 98.0 &96.1 & -\\
        LSTM~\cite{bai2018empirical} & 70k & 87.2 & 85.7 & -\\
        GRU~\cite{bai2018empirical} & 70k & 96.2 & 87.3 & -\\
        TCN~\cite{bai2018empirical} & 70k & 99.0 &97.2 & -\\
        r-LSTM~\cite{trinh2018learning} & 500k& 98.4 & 95.2 & 72.2\\
        IndRNN~\cite{li2018independently} & 83k & 99.0 & 96.0 & -\\
        TrellisNet~\cite{bai2018trellis} & 8M & 99.20 & 98.13 & 73.42\\
        URLSTM~\cite{gu2020improving} & - & 99.28 & 96.96 & 71.00\\
        HiPPO~\cite{gu2020hippo}& 0.5M & - & 98.30 & -\\
        coRNN~\cite{rusch2020coupled} & 134k & 99.4 & 97.3 & -\\
        CKCNN~\cite{romero2021ckconv} & 98k & 99.31 & 98.00 & 62.25\\
        LSSL~\cite{gu2021combining} & - & 99.53 & 98.76 & 84.65\\
        S4~\cite{gu2021s4} & - & 99.63 & 98.70 & \textbf{91.13}\\
        \hline
        \hline
        FlexTCN~\cite{romero2021flexconv} & 375k & 99.62 & 98.63 & 80.82\\
        Ours  & 373k &  \textbf{99.75} &  \textbf{99.10} & \textbf{84.86}\\
        \hline
    \end{tabular}}
\caption{Sequential image classification results.}
\label{table:seqimage}
\end{table}

\begin{table}
\centering
\resizebox{0.45\textwidth}{!}{
    \begin{tabular}{c c|c|c|c}
        \noalign{\smallskip}\noalign{\smallskip}\hline
        Model & Params. & CT & SC & SC-raw \\
        \hline
        GRU-ODE~\cite{de2019gru} & 89k & 96.2 &44.8 & $\sim$ 10.0\\
        GRU-$\Delta$t~\cite{kidger2020neural} & 89k & 97.8 & 20.0 & $\sim$ 10.0\\ 
        GRU-D~\cite{che2018recurrent} & 89k & 95.9 &23.9 & $\sim$ 10.0\\
        ODE-RNN~\cite{rubanova2019latent} & 89k & 97.1 & 93.2 & $\sim$ 10.0\\
        NCDE~\cite{kidger2020neural} & 89k & 98.8 & 88.5 & $\sim$ 10.0\\
        CKCNN~\cite{romero2021ckconv} & 100k & \textbf{99.53} & 95.27 & 71.66\\
        LSSL~\cite{gu2021combining} & - & - & 93.58 & -\\
        S4~\cite{gu2021s4} & - & - & 93.96 & \textbf{98.32}\\
        \hline
        \hline
        FlexTCN~\cite{romero2021flexconv} & 373k & \textbf{99.53} & 97.67 & 91.73\\
        Ours  & 371k &  \textbf{99.53} &  \textbf{97.71} & \textbf{94.95}\\
        \hline
    \end{tabular}}
\caption{Time-series classification results.}
\label{table:timeseires}
\end{table}

\subsection{Image classification}
% \begin{figure*}[ht]
%   \centering
%   % \includegraphics[width=\textwidth]{figs/kernels.jpg}
%   \includegraphics[width=\textwidth]{figs/coord_kernel3.jpg}
%   \caption{Visualization of kernels. Each row shows the location of points and first 4 kernels of a filter. For ease of visualization, the kernels are first subjected to the absolute value operation and then normalized to a range of [0,1].}
%   \label{fig:kernels}
% \end{figure*}
% \subsubsection{Image classification with continuous kernel}
\textbf{Image classification with the continuous kernel.}
We validated our SMPConv on a 2D image dataset, CIFAR10~\cite{krizhevsky2009learning}, which is dominated by discrete convolutions, to show that the continuous kernel can capture spatial information as well. Similar to the experiments on sequential data, we followed the network design choice of FlexNet~\cite{romero2021flexconv}, where the kernel size is $33 \times 33$. More details are in \cref{appxsec1.1}.

% We have compared SMPConv with standard and MLP-based CNNs in \cref{table:cifar}. Although standard CNNs outperformed continuous convolution approaches, it seems that this higher performance of standard CNNs is due to the excessively higher number of learnable parameters, i.e., standard CNN has parameters 20$\times$ than ours, and further, the network details are also different. Interestingly, our continuous kernel representation model SMPConv obtained competitive performance in smaller parameter settings. This indicates the potential advantage of our proposed model.

As shown in \cref{table:cifar}, our continuous kernel representation model slightly outperforms ResNet, a discrete $3 \times 3$ convolution model, with a less number of parameters. It implies that our model is competitive and promising. Our model also showed better performance than MLP-based counterparts, CKCNN-16 and FlexNet-16, even when the parameters of ours were around $30\%$ lesser. In addition, we have already identified the efficiency of our model in \cref{table:deformable}. These results suggest that our method is more suitable for kernel generation than MLP-based implicit formulations.

% Additionally, our model showed better performance than FlexNet even when the parameters of our model were around 30 $\%$ lesser than that of FlexNet. We examined and found that this performance gain comes from the different representations of the kernel. Unlike FlexNet, our kernel representation is formulated by explicit weights with a distance function. This formulation can exploit the inductive bias of CNN more than MLPs-based counterparts. 

\begin{table}
\centering
    \begin{tabular}{c|c|c}
        \noalign{\smallskip}\noalign{\smallskip}\hline
        Model & Params. & Accuracy \\
        \hline
        ResNet-44~\cite{he2016resnet} & 660k & 92.9 \\
        % ResNet-18 & 11.2M & 94.92 \\
        % \hline
        CKCNN-16~\cite{romero2021flexconv} & 630k & 72.1 \\
        FlexNet-16~\cite{romero2021flexconv} & 670k & 92.2 \\
        Ours  & 490k &  \textbf{93.0} \\
        \hline
    \end{tabular}
\caption{2D image classification on CIFAR10.}
\label{table:cifar}
\end{table}

% \subsubsection{Large scale image classification}
\textbf{Large scale image classification.}
Finally, we tested our SMPConv on a large-scale ImageNet dataset~\cite{deng2009imagenet}, which contains more than one million training images and 50,000 validation images. For such a large dataset, the convolution kernels should be carefully trained to model complex data relationships accurately. Through such an experiment, therefore, we can validate that our $\texttt{SMP}$ can represent a descriptive convolution kernel. 

Firstly, we constructed large-scale variants of SMPConv architecture based on RepLKNet~\cite{replknet}. We replaced its discrete depth-wise separable convolution kernel with our \texttt{SMP}. In general, the larger the data and network, the larger the number of filters. To prevent excessive point position parameters depending on the number of filters, we shared the position of points over filters in large-scale settings. We empirically found that this position sharing has little effect on classification performance. 

We proposed two variants of our model, SMPConv-T and SMPConv-B. Thanks to our efficient large kernel, we adjusted the number of channels and blocks so that our variants have a similar number of parameters to the previous models. The number of blocks and channels for each stage is [2, 2, 8, 2] and [96, 192, 384, 768] for SMPConv-T and [2, 2, 20, 2] and [128, 256, 512, 1024] for SMPConv-B, respectively. In RepLKNet-31B, the number of blocks and channels for each stage is [2, 2, 18, 2] and [128, 256, 512, 1024]. More experimental details are provided in \cref{appxsec1.2}.

As reported in \cref{table:imagenet}, our models obtained competitive performance with fewer parameters than existing models. These results show that our kernel representation is promising for large-scale domains as well. Overall, our kernel representation is highly effective and descriptive.

\begin{table}
\centering
\resizebox{0.45\textwidth}{!}{
    \begin{tabular}{c|c|c|c}
        \noalign{\smallskip}\noalign{\smallskip}\hline
        Model & Params. & FLOPs & Top-1 Accuracy \\
        \hline
        ResNet-50~\cite{he2016resnet} & 26M & 4.1G & 76.5 \\
        ResNext-50-32x4d~\cite{xie2017aggregated} & 25M &4.3G & 77.6 \\
        ResMLP-S24~\cite{touvron2022resmlp} & 30M & 6.0G & 79.4 \\
        DeiT-S~\cite{touvron2021deit} & 22M & 4.6G & 79.8 \\
        Swin-T~\cite{liu2021swin} &  28M & 4.5G & 81.3 \\
        TNT-S~\cite{han2021transformer} &  24M & 5.2G & 81.3 \\
        ConvNeXt-T~\cite{liu2022convnet} & 29M & 4.5G & 82.1 \\
        SLaK-T~\cite{slak} & 30M & 5.0G & \textbf{82.5} \\
        SMPConv-T(ours) & 27M& 5.7G & \textbf{82.5}\\
        \hline
        \hline
        DeiT-Base/16~\cite{touvron2021deit} & 87M & 17.6G & 81.8 \\
        Swin-B~\cite{liu2021swin} &  88M & 15.4G & 83.5 \\
        ConvNeXt-B~\cite{liu2022convnet} &  89M & 15.4G & 83.8 \\
        SLaK-B~\cite{slak} &  95M & 17.1G & \textbf{84.0} \\
        \hline
        RepLKNet-31B~\cite{replknet} &  79M & 15.3G & 83.5 \\
        SMPConv-B(ours) &  80M & 16.6G & \textbf{83.8} \\
        \hline
    \end{tabular}}
\caption{2D image classification on ImageNet-1K.}
\label{table:imagenet}
\end{table}

\subsection{Ablation}
\begin{table}
\centering
    \begin{tabular}{c|c|c|c}
        \noalign{\smallskip}\noalign{\smallskip}\hline
        Models & radius & coordinate & Accuracy \\
        \hline
        A &- & - & 90.92 \\
        B &\checkmark & - & 91.35 \\
        C & - & \checkmark & 92.47 \\
        SMPConv & \checkmark & \checkmark &  \textbf{93.00} \\
        \hline
    \end{tabular}
\caption{Ablation study on CIFAR10. A checkmark means that the component is a learnable. In case of SMPConv, for instance, both radius and coordinate are learnable parameters.}
\label{table:ablation}
\end{table}

\begin{table}[t]
\centering
    \begin{tabular}{c|c|c|c|c}
        \noalign{\smallskip}\noalign{\smallskip}\hline
         $\sigma$ & 0.05 & 0.2 & 0.3 & 0.5  \\
        \hline
        Accuracy & \textbf{93.00} & 92.24 & 91.84 & 91.51 \\
        \hline
    \end{tabular}
\caption{Classification results on CIFAR10 with different standard deviation $\sigma$ of point location sampling distribution.}
\label{table:ablationstd}
\end{table}

\begin{table}
\centering
    \begin{tabular}{c|c|c|c|c}
        \noalign{\smallskip}\noalign{\smallskip}\hline
         $r$ & 0.12 & 0.18 & 0.24 & 0.3 \\
        \hline
        Accuracy & \textbf{93.00} & 92.36 & 92.06 & 91.76 \\
        \hline
    \end{tabular}
\caption{Classification results on CIFAR10 with different initial radius $r$.}
\label{table:ablationradius}
\end{table}

\begin{table}[t]
\centering
    \begin{tabular}{c|c|c|c|c|c}
        \noalign{\smallskip}\noalign{\smallskip}\hline
         $N_p$ & 4 & 8 & 16 & 32 & 64 \\
        \hline
        Params. & 250k & 330k & 490k & 809k & 1447k \\
        \hline
        Accuracy & 92.56 & 92.28 & \textbf{93.00} & 92.84 & 92.21\\
        \hline
    \end{tabular}
\caption{Classification results on CIFAR10 with different number of moving points $N_p$.}
\label{table:ablationpoints}
\end{table}

\begin{figure*}[ht]
  \centering
  \includegraphics[width=\textwidth]{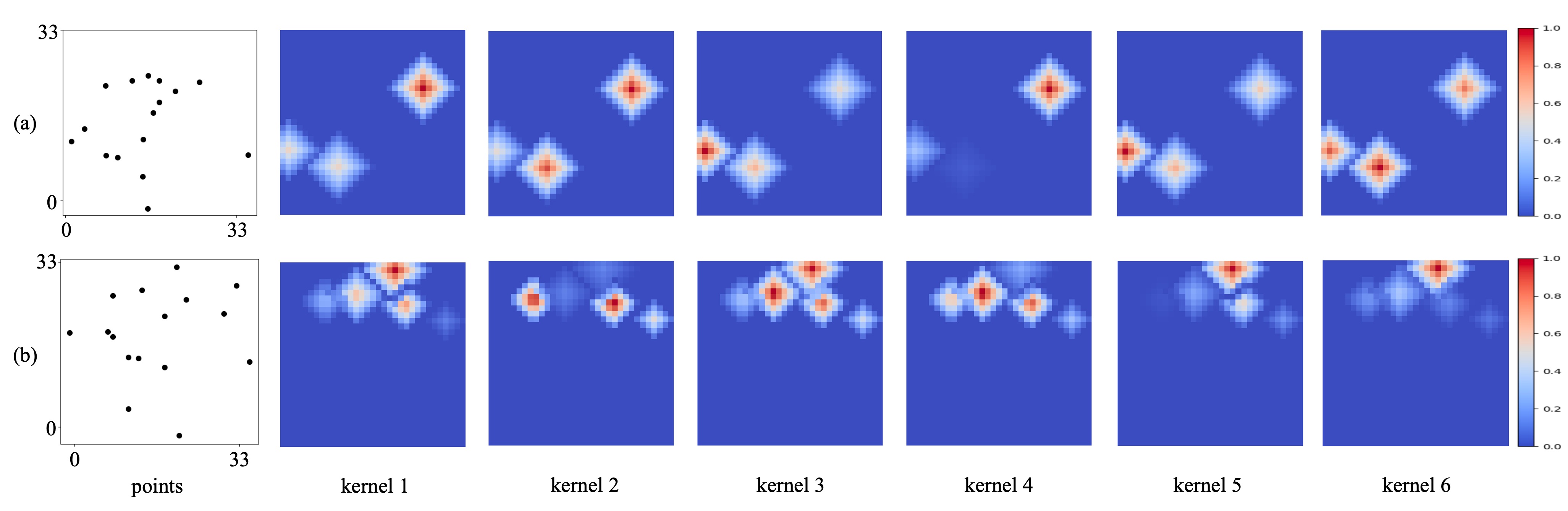}
  \caption{Visualization of kernels. Each row shows the location of points and first 6 kernels of a filter. For ease of visualization, the kernels are first subjected to the absolute value operation and then normalized to a range of [0,1].}
  \label{fig:kernels}
\end{figure*}

We performed various ablation studies with additional experiments on CIFAR10 image classification. First, we investigated the validity of learnable radius and coordinate. In \cref{table:ablation}, it showed that performance degradation occurs when either one or both components are set to non-learnable parameters. Remarkably, Model C, which set coordinate to trainable parameters, outperformed Model A by a considerable margin. Furthermore, Model B also had a slight performance gain. It suggests that even randomly distributed fixed weight points can increase their interpolation ability with trainable radius. These results indicate that training both coordinate and radius is valid.

Next, we identified the effect of the initial position of the points by varying the $\sigma$, a standard deviation of point location sampling distribution. In \cref{table:ablationstd}, we observed that a small $\sigma$ value, indicating initial positions of the points are gathered in the center of the kernel, leads to higher accuracy. This is because it is difficult to train large kernels from the beginning of training. Thus, large kernels can be effectively trained by starting with small kernels and expanding the receptive fields through moving points.

We also found that larger initial radius degrades the model performance as shown in \cref{table:ablationradius}. The large radius results in a large initial kernel size, which makes initial training difficult. Furthermore, it is also challenging to train a large area of the kernel dependent on a single weight point which is not optimized in the early stages of training. Both \cref{table:ablationstd} and \cref{table:ablationradius} empirically show that our initialization methods for $\texttt{SMP}$ are effective.

As depicted in \cref{table:ablationpoints}, we figured out that simply increasing the number of weight points $N_p$ does not helpful for performance. It implies that a small number of points are enough to represent a proper convolution filter. Since the performance is influenced by the number of points, our method is also required tuning like common neural networks. However, the performance difference between CKCNN and FlexNet in \cref{table:cifar} shows that MLP-based methods are severely influenced by architectural settings. That is, they have extensive search space, such as depth, width, and activation function, so they typically require more tuning than our method. Moreover, the impact of the number of points is not particularly significant in that even the worst ($N_p=64$) slightly outperforms the FlexNet (acc=92.20).

\subsection{Visualization}
% \begin{figure*}[ht]
%   \centering
%   % \includegraphics[width=\textwidth]{figs/kernels.jpg}
%   \includegraphics[width=\textwidth]{figs/coord_kernel3.jpg}
%   \caption{Visualization of kernels. Each row shows the location of points and first 4 kernels of a filter. For ease of visualization, the kernels are first subjected to the absolute value operation and then normalized to a range of [0,1].}
%   \label{fig:kernels}
% \end{figure*}

\begin{figure}[t]
  \centering
  \includegraphics[width=0.45\textwidth]{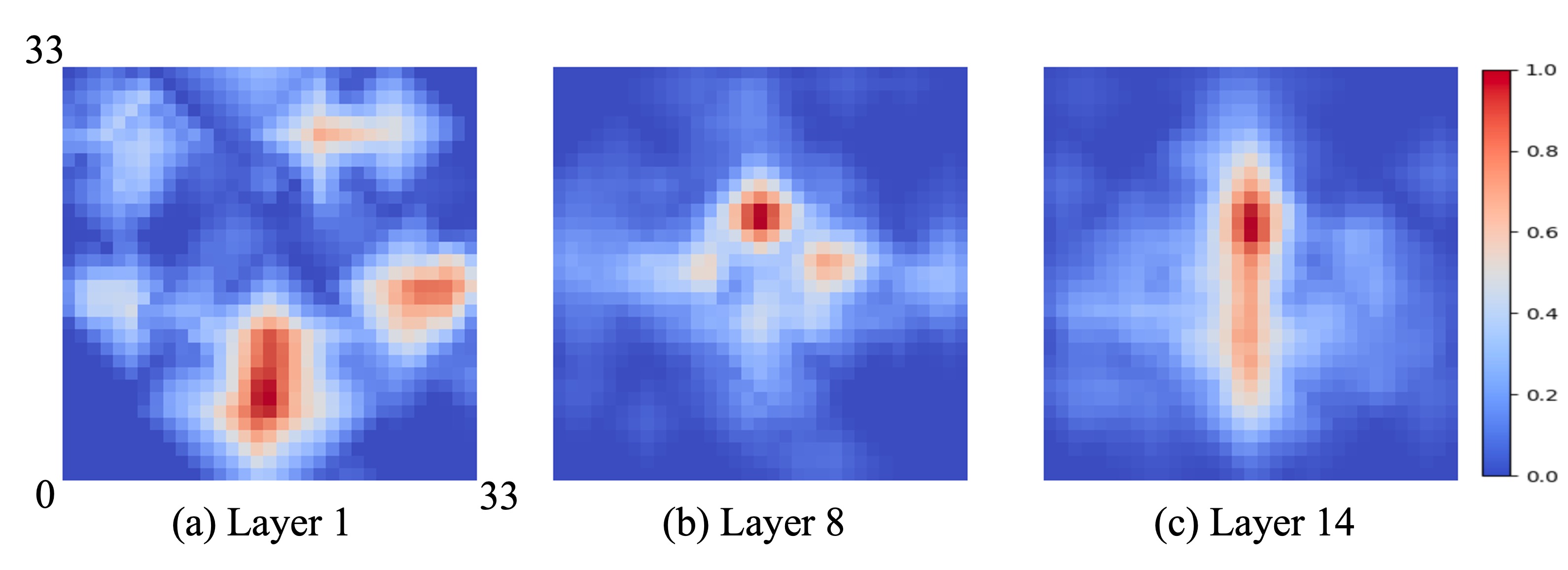}
  \caption{Normalized sum of the absolute value of trained filters. (a), (b), and (c) are top, middle, and bottom layers, respectively.}
  \label{fig:receptive}
\end{figure}

Finally, we analyze our \texttt{SMP} by visualizing filters trained on CIFAR10. In the first column of \cref{fig:kernels}, we can observe the trained weight points' position. In our method, point locations ${p_i}$ are mainly sampled near the center of the kernel for stable training.  It shows that the points spread out for optimal kernel representation over the training process, as we argued, and thus the receptive fields are not limited to a small part. Also, we can figure out that there are square patterns caused by \cref{eq:distance_function} in the kernels, where each square has its own area. This suggests that although the radius parameters are initialized with small values, the values are individually increased and optimized for each corresponding weight point during training.

Observing visualized convolution kernel in \cref{fig:kernels}, kernels from the same filter share the receptive fields. It allows a single filter to focus on the shared area. Furthermore, as illustrated in \cref{fig:receptive}, SMPConv has large adaptive receptive fields which are not conventional square or rectangular shapes. This is because it consists of optimized filters with their own small and large receptive fields. Thus, our method can handle not only global information but also local details.

\section{Conclusion and discussion}

In this paper, we present a method to build a continuous convolution. We propose using point representations, where each point has the weight parameters, coordinates, and radius to learn. By connecting the points, we can implement a continuous function, which can be utilized to construct convolutional kernels. We have provided extensive experimental results, showing that drop-in replacement in the existing training pipeline without bells and whistles improved the performance by a safe margin. We also show that a continuous convolution can be effectively utilized in a large-scale experiment. We expect more research and development in this direction.

Although promising, there are many rooms to be improved. Due to the limited computational budget, we could not conduct sufficient experiments in the large-scale experiment. The experimental results provided in this manuscript resulted from a few runs. As trial and error are essential in the machine learning development process, we plan to find optimal configurations and training techniques to enhance the performance of the proposed method. 

We also observed that the learned kernels often show sparse patterns depending on the tasks. It is well aligned with the success of dilated convolution or its variants, and our methods automatically learn proper sparsity during the training process on specific tasks. Adding prior knowledge through training and regularization techniques would further improve performance, especially for tasks requiring longer-term dependency modeling.

%The points representation in the proposed methods can freely move via coordinate optimizations. However, we observe that a few points often stay still or move beyond the input domain boundary, resulting in not being used when constructing the kernels. This phenomenon would waste the capacity of the parameters. We plan to explore the methods to either prune or restrict its behavior, and we believe fully utilizing the point representations would enhance the performance.

\section*{Acknowledgments}
We thank Usman Ali for valuable discussions. This research was supported by the Ministry of Science and ICT (MSIT) of Korea, under the National Research Foundation (NRF) grant (2022R1F1A1064184,
2022R1A4A3033571), Institute of Information and Communication Technology Planning Evaluation (IITP) grants for the AI Graduate School program (IITP-2019-0-00421), and the BK21 FOUR Project.
%%%%%%%%% REFERENCES
{\small
\bibliographystyle{ieee_fullname}
\bibliography{egbib}
}

% \end{document}

\clearpage
\appendix
\renewcommand\thesection{\Alph{section}}
\setcounter{section}{0}

\twocolumn[
\begin{center}
\Large{\bf{SMPConv: Self-moving Point Representations for Continuous Convolution\\ Appendix}}\par\vspace{3ex}
\end{center}]

\section{Experimental details}
\subsection{Sequential data and image classification} \label{appxsec1.1}
In each convolution filter, SMPConv has 30 weight points for 1D and 16 weight points for 2D. For SMPConv1D, we sample the point locations from zero mean truncated gaussian distribution with $\sigma=0.1$. Because of causal convolution, we sample in the range $(-1, 0)$ rather than $(-1, 1)$. For SMPConv2D, we sample the point locations from 2D zero mean truncated gaussian with $\Sigma = [[\sigma_1,0],[0, \sigma_2]]$, where $\sigma_1=\sigma_2=0.05$. We initialize radius as $r \approx \frac{2}{k}d$, where $k$ is kernel size and $d$ is dimension of input (i.e., $d=1$ for 1D, $d=2$ for 2D). In 2D, the kernel size means the width of the kernel. The size of the additional small kernel is $5$ for 1D and $3 \times 3$ for 2D, respectively. Following FlexConv~\cite{romero2021flexconv}, we use batch normalization~\cite{ioffe2015batch} after convolution and skip connection. 
% To fuse batch normalization after training, we do not use bias terms for convolution.

We train our networks using Adam~\cite{kingma2014adam} optimizer. We use a cosine annealing learning rate scheduling with warm-up epochs. The learning rate for radius parameters is set to be $0.1\times$ smaller than the regular learning rate. During the training, the radius range is clipped from 0.0001 to 1.0. More details for each data are shown in \cref{table:hyperparameters}. For sequential data experiments, we train our model with a single NVIDIA A100 GPU. We use a single RTX3090 GPU for CIFAR10 experiments.

% \vspace{.1cm}
\subsection{Image classification on ImageNet-1k} \label{appxsec1.2}
Our large-scale variants of SMPConv networks have the same architecture as RepLKNet~\cite{replknet} except for large kernel convolution, which is replaced by our $\texttt{SMP}$. Like ~\cite{replknet}, we set the kernel size of each stage to [31, 29, 27, 13] and use additional $5 \times 5$ convolution for reparameterization trick. We use $\lfloor \frac{k^2}{4} \rfloor$ weight points for each $\texttt{SMP}$ depth-wise version, which shares weight points over channels, where $k$ is the kernel size of corresponding each block. The point locations and radius are initialized in the same way as ~\cref{appxsec1.1} SMPConv2D with $\sigma_1 = \sigma_2 = 0.2$.

Our models are trained for 300 epochs using AdamW~\cite{loshchilov2017adamw} optimizer. We set the batch size of 2048. The initial learning rate is set to $4 \times 10^{-3}$ with cosine annealing scheduling and 10 warm-up epochs. We use RandAugment~\cite{cubuk2020randaugment} in Timm~\cite{rw2019timm}("rand-m9-mstd0.5-inc1"), Label Smoothing~\cite{szegedy2016rethinking} coefficient of 0.1, Mixup~\cite{zhang2017mixup} with $\alpha=0.8$, Cutmix~\cite{yun2019cutmix} with $\alpha=1.0$, Rand Erasing~\cite{zhong2020random} with probability of $25\%$, Stochastic Depth with drop path rate of $10\%$ for SMPConv-T, and $50\%$ for SMPConv-B, and model EMA(exponential moving average) with a decay factor of 0.9999. For fast depth-wise convolution computation, we use block-wise(inverse) \textit{implicit gemm} algorithm implemented by ~\cite{replknet}. We train both SMPConv-T and SMPConv-B with 4 NVIDIA A100 GPUs.

\begin{table*}[t]
\centering
    \begin{tabular}{p{2cm}|p{1.5cm}|p{1.5cm}|p{1.5cm}|p{1.5cm}|p{1.5cm}|p{1.5cm}|p{1.5cm}}
        \noalign{\smallskip}\noalign{\smallskip}\hline
          & sMNIST & pMNIST & sCIFAR10 & CT & SC & SC-raw & CIFAR10 \\
        \hline
        lr       & 0.0001 & 0.0001 & 0.0002 & 0.0001 & 0.002 & 0.001 & 0.005 \\
        % \hline
        epoch       & 200 & 200 & 200 & 300 & 300 & 160 & 210 \\
        warm-up      & 5 & 5 & 5 & 5 & 5 & 10 & 10 \\
        % \hline
        dropout  & 0 & 0 & 0 & 0 & 0.2 & 0.1 & 0.1 \\
        % \hline
        \# of batch  & 64 & 64 & 64 & 64 & 32 & 32 & 64 \\
        % \hline
        weight decay  & 1e-5 & 1e-5 & 1e-5 & 1e-5 & 1e-5 & 1e-5 & 1e-5 \\
        % \hline
        kernel size  & 784 & 784 & 1024 & 182 & 161 & 16000 & 33 $\times$ 33 \\
        \hline
    \end{tabular}
\caption{Hyper-parameter details}
\label{table:hyperparameters}
\end{table*}
\section{Additional results}
\subsection{Larger kernels}
We set the kernel size of each stage to [31, 29, 27, 13] for large-scale variants of SMPConv networks following RepLKNet~\cite{replknet}. Although the current kernel sizes are larger than conventional convolution, we evaluate whether our model is trained without performance degradation even when using larger kernels. 

To conduct this experiment, we design a new variant,  SMPConv-mobile. For the mobile variant, the number of blocks and the number of channels for each stage is [2, 2, 2, 2] and [64, 128, 256, 320], respectively. Also, we use $\lfloor \frac{k^2}{8} \rfloor$ weight points for each $\texttt{SMP}$ and reduce the expansion ratio of feed-forward networks from 4 to 2. We train this variant for 120 epochs and do not use Stochastic Depth. Other training settings are same as ~\cref{appxsec1.2}. We set the kernel size of each stage to [31, 29, 27, 13] for SMPConv-mobile31 and [51, 49, 47, 13] for SMPConv-mobile51. 

In ImageNet-1k~\cite{deng2009imagenet} image classification, SMPConv-mobile31 and SMPConv-mobile51 get \textbf{73.5}$\%$ and \textbf{73.7}$\%$ top-1 accuracy, respectively. Thus, using our $\texttt{SMP}$, convolution kernel sizes can be increased without performance degradation, even in large-scale data.

% \begin{table*}[t]
% \centering
%     \begin{tabular}{p{2cm}|p{1.5cm}|p{1.5cm}|p{1.5cm}|p{1.5cm}|p{1.5cm}|p{1.5cm}|p{1.5cm}}
%         \noalign{\smallskip}\noalign{\smallskip}\hline
%           & sMNIST & pMNIST & sCIFAR10 & CT & SC & SC-raw & CIFAR10 \\
%         \hline
%         lr       & 0.0001 & 0.0001 & 0.0002 & 0.0001 & 0.001 & 0.001 & 0.005 \\
%         % \hline
%         epoch       & 200 & 200 & 200 & 300 & 300 & 160 & 210 \\
%         warm-up      & 5 & 5 & 5 & 5 & 5 & 10 & 10 \\
%         % \hline
%         dropout  & 0 & 0 & 0 & 0 & 0.2 & 0.1 & 0.1 \\
%         % \hline
%         \# of batch  & 64 & 64 & 64 & 64 & 64 & 64 & 64 \\
%         % \hline
%         weight decay  & 1e-5 & 1e-5 & 1e-5 & 1e-5 & 1e-5 & 1e-5 & 1e-5 \\
%         % \hline
%         kernel size  & 784 & 784 & 1024 & 182 & 101 & 16000 & 32 $\times$ 32 \\
%         % \hline
%     \end{tabular}
% \caption{Hyper-parameter details}
% \label{table:hyperparameters}
% \end{table*}
%%%%%%%%% REFERENCES
% {\small
% \bibliographystyle{ieee_fullname}
% \bibliography{egbib}
% }

\end{document}